\documentclass[sigconf]{acmart}
\AtBeginDocument{%
  \providecommand\BibTeX{{%
    \normalfont B\kern-0.5em{\scshape i\kern-0.25em b}\kern-0.8em\TeX}}}

\copyrightyear{2023}
\acmYear{2023}
\setcopyright{rightsretained}
\acmConference[HRI '23]{Proceedings of the 2023 ACM/IEEE International
Conference on Human-Robot Interaction}{March 13--16,
2023}{Stockholm, Sweden}
\acmBooktitle{Proceedings of the 2023 ACM/IEEE International
Conference on Human-Robot Interaction (HRI '23), March 13--16, 2023,
Stockholm, Sweden}
\acmDOI{10.1145/3568162.3578630}
\acmISBN{978-1-4503-9964-7/23/03}

\usepackage{array}
\usepackage{color,soul}
\begin{document}

\title{Exploring Levels of Control \\ 
for a Navigation Assistant for Blind Travelers}
\renewcommand{\shorttitle}{Exploring Levels of Control for a Navigation Assistant for Blind Travelers}
\author{Vinitha Ranganeni}
\affiliation{%
  \institution{University of Washington}
  \city{Seattle}
  \state{Washington}
  \country{USA}
}
\email{vinitha@cs.washington.edu}

\author{Mike Sinclair}
\affiliation{%
  \institution{Microsoft Research}
  \city{Redmond}
  \state{Washington}
  \country{USA}
}
\email{sinclair@microsoft.com}

\author{Eyal Ofek}
\affiliation{%
  \institution{Microsoft Research}
  \city{Redmond}
  \state{Washington}
  \country{USA}
}
\email{eyalofek@microsoft.com}

\author{Amos Miller}
\affiliation{%
  \institution{Independent Researcher}
  \city{Seattle}
  \state{Washington}
  \country{USA}
}
\email{miller.amos@gmail.com}

\author{Jonathan Campbell}
\affiliation{%
  \institution{Microsoft Research}
  \city{Redmond}
  \state{Washington}
  \country{USA}
}
\email{jon.campbell@microsoft.com}

\author{Andrey Kolobov}
\affiliation{%
  \institution{Microsoft Research}
  \city{Redmond}
  \state{Washington}
  \country{USA}
}
\email{akolobov@microsoft.com}

\author{Edward Cutrell}
\affiliation{%
  \institution{Microsoft Research}
  \city{Redmond}
  \state{Washington}
  \country{USA}
}
\email{cutrell@microsoft.com}

\renewcommand{\shortauthors}{Ranganeni et al.}
\newcommand{\figref}[1]{Fig.~\ref{#1}} 
\newcommand{\subfigref}[2]{\figref{#1}\subref*{#2}} 
\begin{abstract}
Only a small percentage of blind and low-vision people use traditional mobility aids such as a cane or a guide dog. Various
assistive technologies have been proposed to address the limitations of traditional mobility aids. These devices
often give either the user or the device majority of the control. In this work, we explore how varying levels of control affect the users’ sense of agency, trust in the device, confidence, and successful navigation. We present Glide, a novel
mobility aid with two modes for control: Glide-directed and User-directed. We employ Glide in a study (N=9) in which blind
or low-vision participants used both modes to navigate through an indoor environment. Overall, participants found that Glide was
easy to use and learn. Most participants trusted Glide despite its current limitations, and their confidence and performance increased as they continued to use Glide. Users’ control mode preferences varied in different situations; no single mode ``won'' in all situations.

\end{abstract}



\begin{teaserfigure}
  \includegraphics[width=\textwidth]{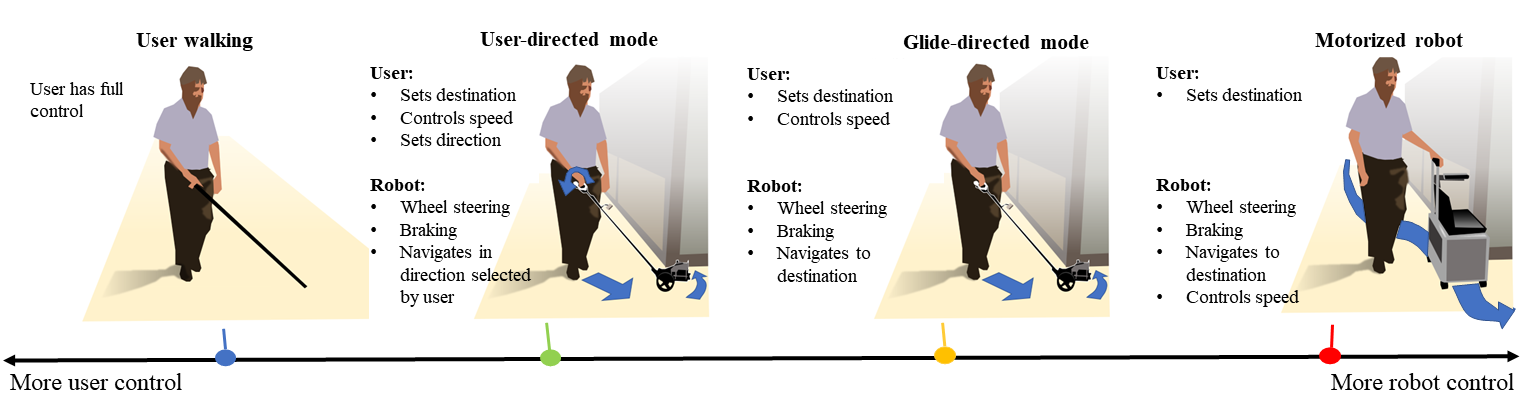}
  \Description[Different mobility aids for blind and low-vision people]{This figure shows a spectrum of user control for different guiding devices. The far-left figure shows a user with a cane. The middle-left and middle-right figure shows a user being guided by Glide in the user-directed and glide-directed mode respectively. The right most figure shows a user being guided by a motorized robot.}
  \caption{ 
  Different mobility aids for blind and low-vision people vary in the autonomy and navigational control they offer to the user. White canes (left) give the user full control of their movement, but short-range sensing making it difficult to navigate in unfamiliar areas. A motorized robot (far right) may follow a global map to a target, but leaves little control to the user as it dictates the path and the walking speed. We explore two different mixed-control schemes, where the user pushes the proposed {\em Glide} navigation assistant in front of her. In the {\em User-directed} mode (middle-left), the user sets the destination and Glide steers the user's walking direction. In the {\em Glide-directed} mode (middle-right), the user is notified of existence of possible turns at junctions and have additional control on the the chosen path. 
  }
  \label{fig:teaser}
\end{teaserfigure}


\maketitle

\begin{figure}[t!]
  \includegraphics[width=\columnwidth]{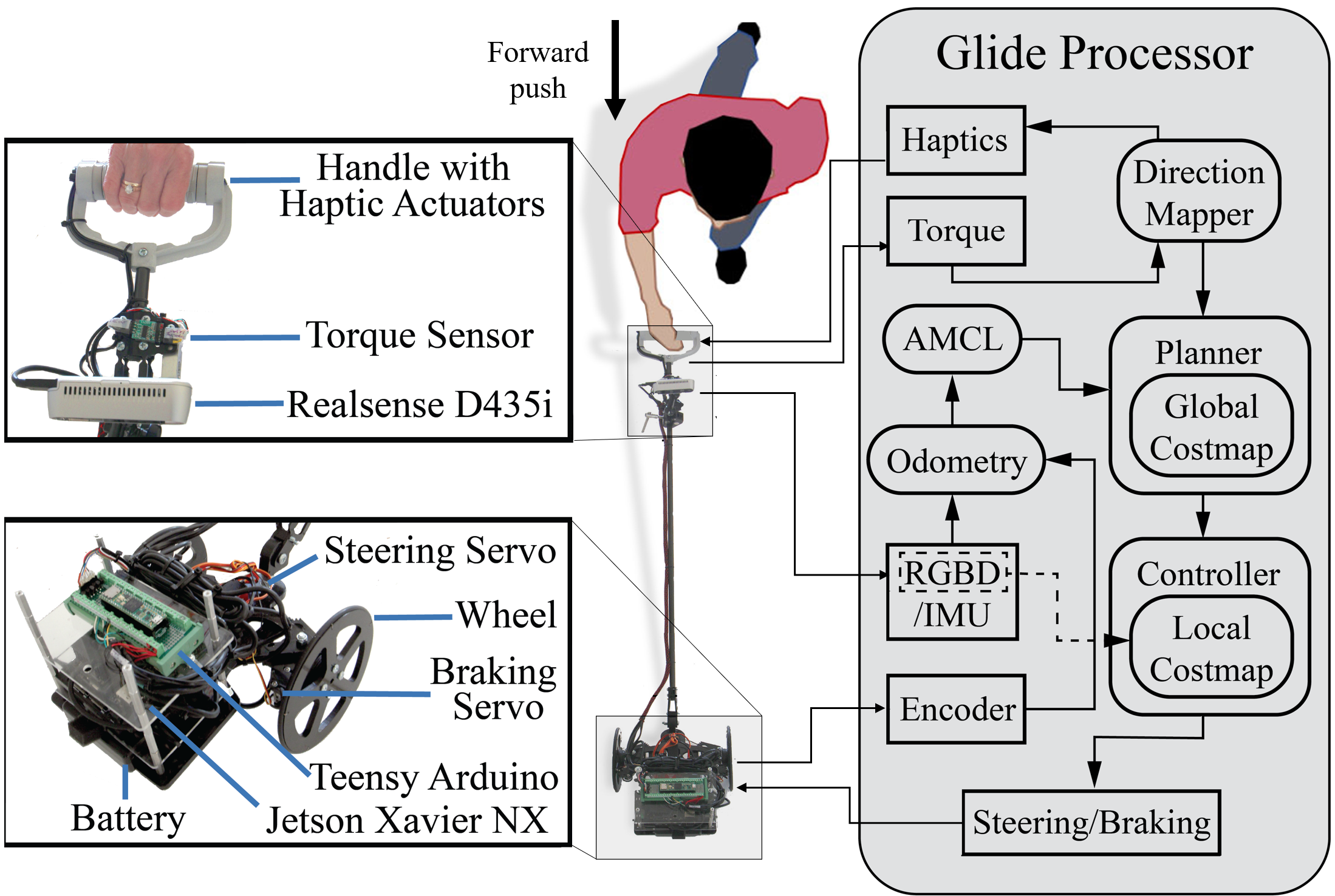}
  \Description[A breakdown of the components of Glide and its processor]{The figure shows a picture of Glide and zooms into the base and handle. The various sensors and actuators are labeled. On the base there are steering and braking servos, two wheels, a teensy arduino, battery and jetson xavier NX. There is a pole that connects the base to the handle. The handle has haptic actuators. Below the handle there is a torque sensor and a Realsense D435i. The right side shows how the Glide processor works. The torque to converted into a direction mapping which determines the haptics and is passed to the planner. The RGBD, IMU and encoder data is passed to the odometry and ACML package for localization. The planner takes the localization information and direction mapping and generates a plan sent to the controller which then determines the steering angle and whether to engage/disengage the breaks. Additionally, the RGBD data is used to generate a local costmap.}
  \caption{A breakdown of the components of Glide and its processor. The user can twist the handle to indicate desired direction of travel (input to torque sensor) and pushes the robot forward (input to encoder). Glide outputs haptic feedback, steers the user and applies/releases the brakes.}
  \label{fig:glide}
\end{figure} 

\section{Introduction}
As of 2016, approximately 7 million people in the US reported being blind or having low vision (BLVI)~\cite{statisticsnational}. For these people, white canes and guide dogs are the only primary mobility aids. White canes come in direct contact with the area immediately ahead of their user, enabling the user to sense the environment, assess differences in materials, and detect obstacles. Guide dogs lead the user around obstacles and are capable of global navigation in familiar settings. The user can interact with the dog and suggest different walking directions. Despite empowering many people, these mobility aids require long training to master and use confidently, and even then the risk of losing one’s way remains substantial. The difficulty of safe navigation can also adversely impact a person’s \emph{self}-confidence. As a result, of the estimated 7 million BLVI in the US, \textbf{only 2\% to 8\% use a white cane}~\cite{statisticsnational}. \textbf{Only about 2\% of the remaining individuals use guide dogs}~\cite{statisticsnational}.  About 90\% of BLVI have a high dependency on sighted assistance and/or confine their lives to a limited set of locations and activities, in many cases solely to their homes. Prior efforts to automate blind navigation, such as smartphone-based navigation applications and motorized robots, so far have failed to change this reality.

In this work, we conjecture that a mobility aid's helpfulness depends on a heretofore understudied factor --- the level of control it offers to its user~(\figref{fig:teaser}). White canes and smartphone navigation programs leave full navigational control to the person, but the amount of information they give to the user may not be enough for local decision making. Guide dogs and motorized robots are closer to the other end of the control spectrum, but a BLVI person following them may feel a loss of agency in the process. To experiment with different levels of shared control, we developed \textbf{Glide}~(\figref{fig:glide}), a novel mobility aid designed to safely steer users to their destination with a range of a haptic vocabulary that conveys a tactile sense of the surface it rolls on while enabling the user to set the walking pace. Glide is designed to be light and portable. Beside the practicality of users carrying it over stairs or bringing it with them, users can manipulate the device to their liking to increase the sense of control.

Glide uses passive kinetic guidance through a pole with a handle connected to a small mobile platform with steerable and brake-able wheels~(\figref{fig:glide}). The wheels are non-motorized and require the user to push the device in front of them. Glide’s sensors allow it to identify the user’s location and both static and dynamic obstacles. Glide uses this information to guide the user around obstacles or engage the brakes to make the user stop. 
Glide’s handle connects the user’s grip to the wheeled platform, conveying the ground's tactile information. The handle is equipped with an array of haptic actuators that serve as another channel of
communication with the user (e.g., to slow down when approaching the goal). This enhances the user’s understanding of their surroundings and helps them build a mental map of their environment.


\begin{figure*}[t]
  \includegraphics[width=\textwidth]{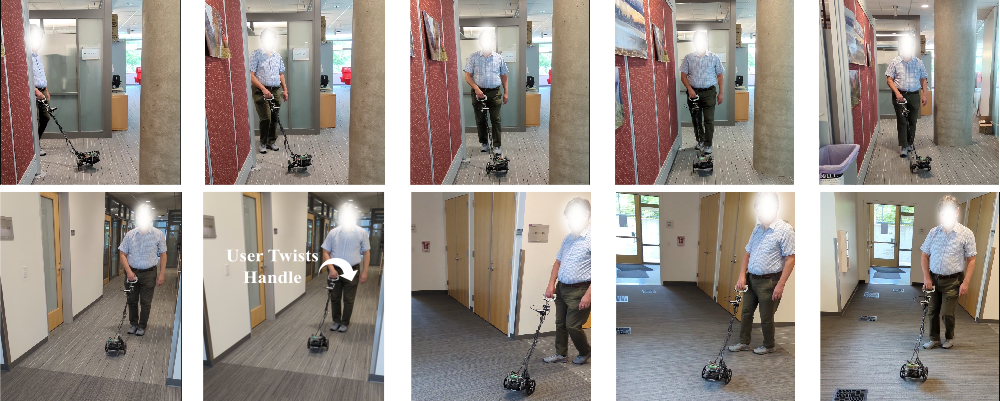}
  \Description[Floorplans showing the routes for each mode]{The left figure shows the floor plan used in the study. The route used for the glide-directed mode is drawn on the floor plan. The route includes, in order, two 90-degree left turns, a right turn that includes avoiding a pillar, followed by another right turn. The right figure also shows a floor plan used in the study for the user-directed mode. There are four routes drawn on the floor plan that lead to three different destinations. One route leads to a lounge, one route leads to a kitchen and two routes lead to a work area. The routes all start in the same position and consist of various turns and obstacles that need to be avoided to get to their respective goals.}
  \caption{(Top) The Glide-directed mode shows the user being guided around a corner while avoiding a pillar. (Bottom) In the User-directed mode the user twists the handle to the left and Glide guides the user along a left turn.}
  \label{fig:glide_demo}
\end{figure*}

We experimented with two modes of operation of Glide, using increasing levels autonomy~(\figref{fig:glide_demo}). In the \textbf{Glide-directed} mode, the user 
pushes Glide forward, while 
Glide steers their walking direction to their desired destination while avoiding obstacles. In the \textbf{User-directed} mode, Glide waits for the user’s directional input at decisions points, such as junctions in a hallway, and then steers them in their desired direction of travel to the next decision point while avoiding obstacles. We tested Glide in an indoor office building, where turns are mostly limited to 90 degrees and straight corridors between them. 

We conducted a user study with nine BLVI people to evaluate the users’ progression through the two control modes and the overall user experience. More specifically, we wanted to understand the
users’ level of trust in Glide, their level of confidence when using Glide, Glide’s ease of use, learnability and whether users’ performance improved as they used Glide. 

Our findings show that Glide was easy to learn, and users found both modes easy to use. Most participants trusted Glide to
avoid obstacles but limitations in the system’s current ability to react quickly impacted other participants’ trust in the device. Overall, most users were confident when using Glide and their confidence level increased as they continued to use the device. Additionally, users’ preferences for modes varied. Most users’ preference for a mode was situational. Few users stated that they strongly preferred one mode over the other.


\section{Related Works}
We discuss Orientation \& Mobility training (O\&M), conventional mobility aids and their limitations. We then discuss various
assistive technologies that have been developed to address the limitations of traditional mobility aids. 

\subsection{Orientation \& Mobility Training}
The purpose of Orientation \& Mobility training (O\&M) is to teach people with visual impairments how to travel safely through a variety of environments~\cite{jacobson1993art}. This training helps people learn how to use other senses to gain new information about their surroundings and navigate safely. Some techniques taught include
maintaining a straight line of travel and classifying objects into obstacles, clues, landmarks, or hazards. For a more detailed discussion on O\&M training, see~\cite{jacobson1993art, mettler1998cognitive, swobodzinski2009indoor}.

O\&M skills are used in conjunction with primary mobility aids such as a white cane or guide dog, each of which requires substantial training to use effectively. Typically, users must be trained for more than 100 hours to become skilled with the white cane. Effective use of a guide dog depends on the user being competent with a cane and having established O\&M skills. It can take up to 6 months to start working with a new guide dog, and their working life is typically 6 to 7 years.

There are additional limitations, mostly due to the local nature of the aids. While a cane can be used to sense the immediate vicinity of a user, it relies on the user’s familiarity with the environment to navigate to a destination. A dog can help guide the user to a limited number of known locations while avoiding obstacles.

\begin{figure*}[t]
\centering
  \includegraphics[width=0.9\textwidth]{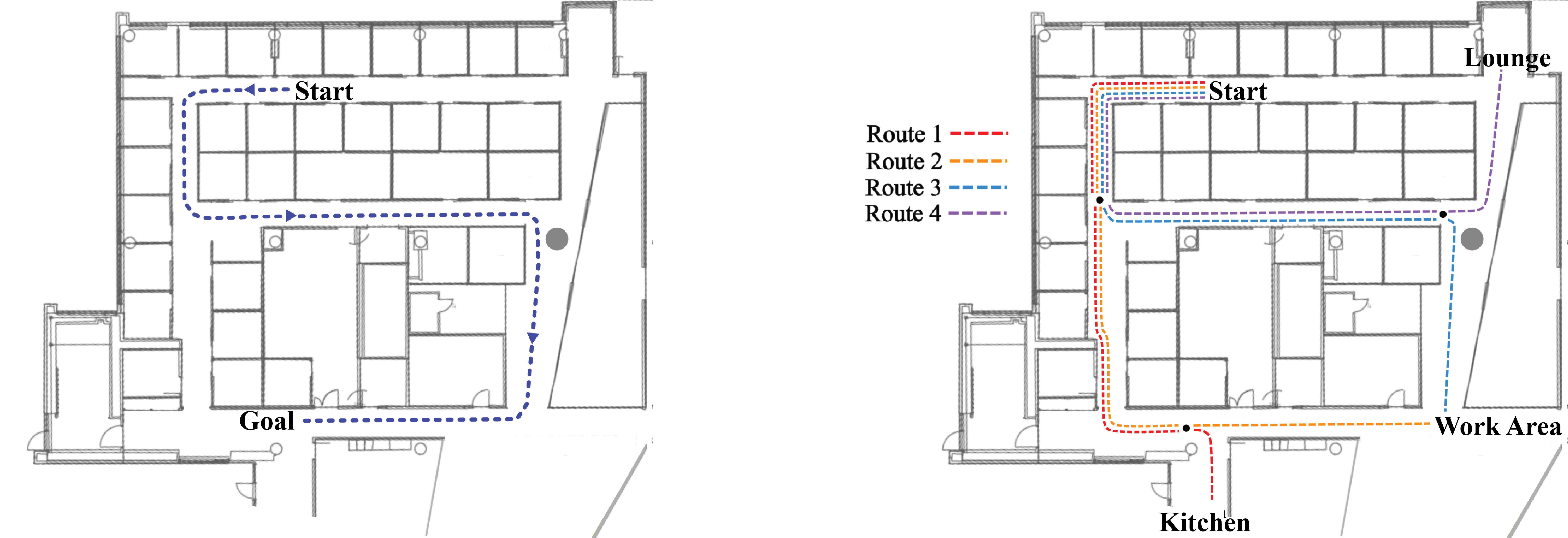}
  \Description[This figure shows five consecutive frames of Glide guiding a user in the glide-directed and user-directed modes.]{This figure shows five consecutive frames of Glide guiding a user in the glide-directed and user-directed modes. There are two rows. The top row shows the user being guided by Glide in the glide-directed mode. Glide guides the user around a corner and a pillar. The bottom row shows the user being guided by Glide in the user-directed mode. The user reaches a T-junction and twists the handle left. Glide then guides the user to the left.}
  \caption{Explored shared-control schemes. Glide-directed mode: (left) The user walks, pushing Glide, and their walking direction is set by Glide to follow the blue path to the goal point. User-directed mode: (right) The user is asked to provide Glide a direction at decision points (colored paths shows options available to the user).}
  \label{fig:routes}
\end{figure*}

\subsection{Assistive Technologies for Blind Users}

Over the years researchers have developed a variety of assistive technologies to aid people with visual impairments by detecting and avoiding obstacles, improving orientation and virtual wayfinding~\cite{roentgen2008inventory}. Explorations have included many diverse types of devices with varying levels of autonomy. 

The most readily available assistive technology supplement O\&M skills. This includes commercial smartphone applications such as Google Maps~\footnote{https://www.google.com/maps/preview}, which visually identifies landmarks, provides directions, and helps the user regain orientation, and Soundscape~\footnote{https://www.microsoft.com/en-us/research/product/soundscape/}, which helps the user navigate using spatial audio that enhances their awareness of their surroundings and the direction to their destination. Both applications use GPS for localization and hence are limited to outdoor navigation. Researchers have also presented various types of navigation systems that provide turn-by-turn navigation assistance to help blind users walk to their destination~\cite{ahmetovic2016navcog, fallah2012user, sato2019navcog3, fusco2020indoor}. Most of these systems, however, are unaware of obstacles that were not in the initial map of the environment. To help blind users avoid collisions, traditional mobility aids have been augmented with sensors to detect obstacles with non-contact sensing. WeWalk~\footnote{https://wewalk.io/en/} is a commercial device that attaches to a cane, detects obstacles, and provides GPS navigation. Previous augmented white canes detected obstacles in front of the user~\cite{gallo2010augmented} or at trunk and head level~\cite{pyun2013advanced}. While these devices can alert a pedestrian of obstacles, the user must still avoid obstacles by themselves.

Researchers have proposed replacing traditional mobility aids with wearable devices that alert the user of obstacles and navigate them to their destination using haptic~\cite{wang2017enabling} or audio~\cite{fiannaca2014headlock} feedback. These devices achieve hands-free navigation but require custom interfaces and can be too heavy or cumbersome to wear or hold. Robotic navigational aids can be an alternative to wearable devices. CaBot~\cite{guerreiro2019cabot} is a fully autonomous suitcase-like robot that guides users to a destination. This system, however, lacks shared control with the user, as the user follows a motorized robot’s direction and pace. Kayukawa et. al~\cite{kayukawa2020guiding} developed a similar platform but the user can choose to enable an autonomous mode that navigates them with speech or directional guidance around obstacles. 

\subsection{Navigation Through Intersections}

Navigation through intersections has been previously explored by, e.g., Kuribayashi et. al.~\cite{kuribayashi2022corridor}, who proposed a smartphone-based app that provides an obstacle-avoiding path and intersection detection. This work, however, requires the user to hold both a phone and cane at the same time and may not be suitable for all blind travelers. Lacey et. al.~\cite{lacey2000context} proposed PAM-AID, a “smart walker” that aims to assist the elderly BLVI to walk safely indoors. They use a Bayesian network approach that combines sensor information with user input (three buttons for moving forward, left, or right) which activate autonomous robot control in the desired direction. While this is similar to our User-directed mode, the authors do not explore varying levels of control.

\subsection{Cane-like Navigation Assistants}
GuideCane~\cite{ulrich2001guidecane} and the Robotic Cane~\cite{aigner1999shared}, each propose a pole attached to a mobile robot base with passive wheels, allowing the user to push the device as it steers around obstacles. Both devices do not provide autonomous navigation to a set goal but will guide the user in avoiding obstacles. The user specifies desired walking direction by pressing directional buttons on the cane. 

Augmented Cane~\cite{slade2021multimodal} is a white cane with an omni-wheel at the tip that allows the user to control the forward speed while the device steers. This device provides obstacle avoidance, indoor/outdoor navigation, and object localization, yet it is reported as heavy and does not allow the user to input their desired direction of travel. The Co-Robotic Cane~\cite{ye2016co} has a rolling tip and two operational modes: an active mode that steers the user to the desired direction of travel while avoiding obstacles and a passive mode where the device behaves as a white cane but also provides speech feedback on the desired travel direction and obstacle information. The device detects the human intent and automatically switches between modes but does not provide autonomous indoor navigation.

Inspired by these works we propose Glide navigation assistant: portable and lightweight, with flexible control scheme enabling both autonomous indoor navigation and obstacles avoidance, as well as manual pace and intuitive walking direction control by twisting the handle. We use Glide to explore distinct levels of controls of navigation aids.

\begin{table*}[t]
  \caption{Demographic information of study participants}
  \label{tab:demographic}
  \begin{tabular}{lllp{2.75in}p{1.25in}l}
    \textbf{ID} & \textbf{Age} & \textbf{Gender} & \textbf{Vision Level} & \textbf{Impairment Duration} & \textbf{Primary Mobility Aid(s)}\\
    \midrule
    \textbf{P1} & 40 & Male & No functional vision & >10 years & White cane \\ [-0.7mm]
    \midrule
    \textbf{P2} & 56 & Female & No functional vision; some light perception & >10 years & White cane and guide dog\\ [-0.7mm]
    \midrule
    \textbf{P3} & 65 & Male & No vision & >10 years & White cane and guide dog\\ [-0.7mm]
    \midrule
    \textbf{P4} & 56 & Male & No vision & >10 years & White cane and guide dog\\ [-0.7mm]
    \midrule
    \textbf{P5} & 45 & Male & No vision & >10 years & White cane\\ [-0.7mm]
    \midrule
    \textbf{P6} & 19 & Female & Legally blind; no peripheral vision; poor depth perception; no night vision & >10 years & White cane\\ [-0.7mm]
    \midrule
    \textbf{P7} & 80 & Male & No functional vision & >10 years & Walker\\ [-0.7mm]
    \midrule
    \textbf{P8} & 27 & Female & No functional vision; some light perception & >10 years & White cane\\ [-0.7mm]
    \midrule
    \textbf{P9} & 30 & Female & No vision & 5 years & White cane\\ [-0.7mm]
  \bottomrule
    \end{tabular}
\end{table*}

\section{The Glide System Design}

\subsection{Hardware}
Glide is a robotic device with a user-centered design. We designed a passive robot, pushed by the user, to reduce the physical workload while carrying a sufficient payload. This design also saves a significant amount of power and weight by not having to carry the extra battery power required to power the wheels. Additionally, users' walking pace may rapidly change based on their environmental stimuli. Our goal for this design is to eliminate the sense of being dragged by the device by giving the user control over their movement. 
Another important design choice was making Glide lightweight and portable. This enables users to carry it if needed (e.g. stairs), and  easily pull, rotate and direct it as they wish.

Glide has a pole connected to a small platform, that rolls on the floor, using steerable, brake-able, and unpowered wheels. The user holds the handle connected to the pole~(\figref{fig:glide}). The user pushes Glide in front of them, and can feel, through the handle, the wheels following any irregularities in the terrain. There is also a linear array of six vibrotactile actuators (ERM) in the handle to render intuitive haptic symbols to the user’s hand. In contrast to a cane form factor~\cite{slade2021multimodal}, Glide’s base, roughly 9-by-9 inches in size, rolls on the ground and supports the majority of the weight. The total weight of Glide is approximately 3 pounds.

Glide’s sensors enable it to navigate through the environment and detect obstacles. Glide is equipped with an IMU and wheel encoders for computing odometry, a Realsense D435i camera for sensing obstacles, servos for braking and steering, vibrotactile actuators for providing haptic feedback and a Jetson Xavier NX for onboard processing. Glide also has a Teensy 4.1 Arduino processor for handling a torque sensor that is used for sensing the user’s directional input by twisting the handle and the haptic actuators.

Rendering feedback to the user in the form of simple vibrotactile spatial patterns in the handle can help reduce cognitive load~\cite{martinez2014cognitive}, is robust to noisy environments, and does not load the user’s auditory sensing. There are six vibrotactile actuators spread across the handle with three vibration patterns: the left three actuators vibrate when the user twists the handle to the left, the right three vibrate when the user twists the handle to the right and all the vibrators will actuate to indicate that the user should slow down. Many other intuitive haptic symbols are possible, depending on the situation.

To summarize, the user must push Glide for it to move (input to the encoder) and can input their desired direction of travel by twisting the handle (input to torque sensor). Glide will steer the user (output to steering servo), engage or disengage brakes (binary output to brake servos) and provide haptic feedback to the user (output to vibrotactile actuators).

\subsection{Navigation System Design}
Glide uses ROS2 (Robot Operating System), an open-source platform that provides software for robot control. We specifically use Nav2 within ROS2 for navigation. Glide computes odometry by fusing sensor data from wheel encoders, an IMU and a RGBD camera odometry node. It uses the adaptive Monte Carlo localization (AMCL) package for localization and Regulated Pure Pursuit planner for local planning. We rely on a pre-rendered floor plan of our building as a global map. While Nav2 allows for a global planner to be integrated, for the purpose of the study we generated the global plans offline.
While Glide's navigation software runs in a closed loop system, the inclusion of a human in the loop requires a special design. The device cannot move by itself, and requires the user to push it. If the user inputs their desired direction of travel, the device's navigation system has to change its global plan to take the user's feedback into account (see~\figref{fig:glide}).

\section{Modes of Glide}

\subsection{Glide-Directed}
One skill taught during O\&M training is route planning: learning how to get information about your destination and how to get there. A part of this skill is first building a mental map of your environment. Initial navigation through unfamiliar environments, without a mental map, while using a cane or guide dog can be challenging. The Glide-directed experience intends to alleviate that challenge by guiding a user from their current location to their destination while avoiding obstacles. In this mode, Glide generates a global plan to the goal. As the user pushes Glide, the local controller steers them to follow the global plan and avoid any obstacles. When the user is 2 meters from the goal, six vibrotactile actuators provide feedback to the user to slow down and the brakes engage when the user has reached the goal. Note, this mode is similar to the functionality provided by the Augmented Cane and CaBot. however, our feedback modalities are different.

Note, for the purpose of the study we preset the destination. User goal setting is outside the scope of this work as our goal was to evaluate the effectiveness of Glide steering the user to their destination, not find the optimal interaction modality for goal setting. In future work, we will explore interaction modalities for goal setting.

\subsection{User-Directed}
Another skill taught during O\&M training is independent movement: using landmarks and clues to help the person know where they are along a particular route. This helps people with visual impairments learn new routes. In this mode, we verbally inform the user of which directions are available when they reach a junction (in the future, Glide will provide this information). By providing a description of the junction, 
the user can learn unfamiliar routes and later choose to navigate with their primary mobility aid instead of Glide. This mode is similar to the GuideCane, Robotic Cane and Co-Robotic Cane in that the user has control over their desired direction of travel but different in that it allows users to navigate through indoor environments to specific known destinations.

When the user reaches a junction, Glide will engage the brakes, causing the user to stop. The user can then select one of three directions: left, right, or forward. The user can twist the handle left or right to indicate which direction they would like to turn. We map the values from the torque sensor on the handle to two discrete directions (left and right) during the handle twist. When Glide receives a handle twist command, it will return a haptic notification from the left three vibrotactile actuators for a left twist or the right three for a right twist, acknowledging the user input. After the vibrotactile feedback is provided, the brakes disengage, Glide plans a global path to the next junction. The user can begin walking and the controller will steer them to follow the global plan. There are three types of junctions the user can encounter:
\begin{itemize}
    \item \textbf{T-junction}: Turning left or right are the only options. The brakes will remain engaged until the user twists the handle left or right.
    \item \textbf{L-junction}: Going straight and either left or right are the only options. The user can begin walking forward without an explicit input as Glide defaults to a forward global plan to the next junction (if one exists). If the user twists the handle in a direction where there is no feasible path, the brakes will engage for a fixed duration and then disengage allowing user to walk forward if they would like to.
    \item \textbf{Four-way junction}: The user can go straight, left or right.
\end{itemize}

\begin{table}[t]
  \caption{Trial time (min) across participants for each mode.}
  \label{tab:time}
  \begin{tabular}{llllll}
    \textbf{Mode} & \textbf{Trial} & \textbf{Avg} & \textbf{SD} & \textbf{Min} & \textbf{Max}\\ [-0.7mm]
    \midrule
    \textbf{Glide-directed} & 1 & 4.84 & 3.32 & 3.25 & 6.17 \\ [-0.7mm]
    \midrule
    \textbf{Glide-directed} & 2 & 4.62 & 1.39 & 3.78 & 6.67 \\ [-0.7mm]
    \midrule
    \textbf{Glide-directed} & 3 & 4.37 & 0.89 & 3.01 & 6.17 \\ [-0.7mm]
    \midrule
    \textbf{User-directed} & 1 & 3.18 & 1.09 & 2.27 & 5.0 \\ [-0.7mm]
    \midrule
    \textbf{User-directed} & 2 & 2.95 & 1.19 & 1.97 & 3.57 \\ [-0.7mm]
    \midrule
    \textbf{User-directed} & 3 & 3.15 & 1.07 & 2.52 & 4.0 \\ [-0.7mm]
  \bottomrule
    \end{tabular}
\end{table}

\section{User Study}
We conducted a user study with 9 BLVI participants. The main goals were to 1) understand if users trusted Glide, 2) understand if users were confident when using Glide, 3) evaluate if users’ performance increased over time (i.e., reduction in the number of errors) and 4) understand if Glide was easy-to-use and learn.

\subsection{Participants}
We recruited 9 participants who are blind or low-vision to be part of this study (Table 1). The inclusion criteria to be a participant in this study was to have no functional vision. Most participants described themselves as confident travelers on familiar routes \textit{(Mdn=5, SD=0.48)} and not confident travelers on unfamiliar routes \textit{(Mdn=2, SD=0.66)}, ranging from \textit{Not at all Confident} (1) to \textit{Extremely Confident} (5). We acknowledge that a sample size of 9 is too small for a quantitative study so we conducted a qualitative study that focuses on relaying the unquantifiable experiences of the participants.

\subsection{Study Design}
Our study was approved by our Institutional Review Board (IRB). We first obtained informed consent from all participants, provided an overview of the study, and explained how to operate Glide. We informed the user that an experimenter would always be close by to guarantee their safety and would only intervene if necessary. 

We divided the study into three sections. In the first section our goal was to evaluate the Glide-directed mode, where the user was guided by Glide from a fixed starting position to predefined goal destination. Each user completed the walking course three times~(\figref{fig:routes}). Due to a current limitation of the system, we advised participants to walk at a slow pace. We told them to “walk as though they were placing one foot in front of the other.” 

\begin{figure*}[h!]
  \includegraphics[width=\textwidth]{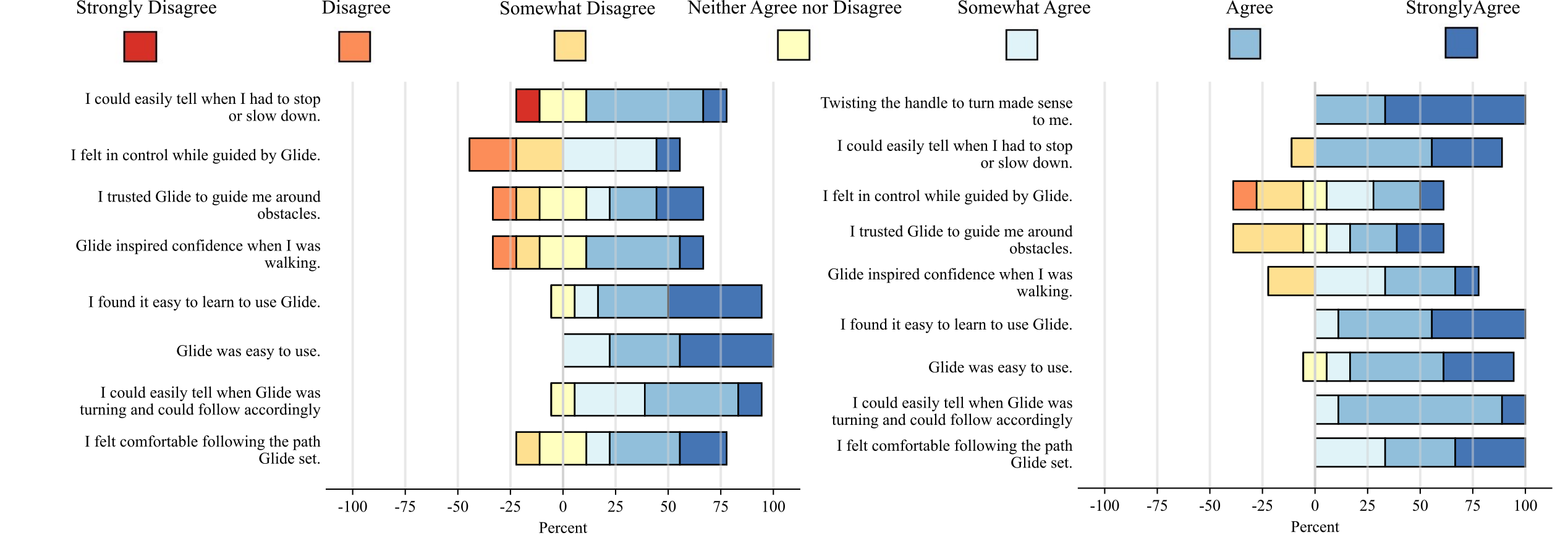}
  \Description[Horizontal box plots showing participants agreement with various statements for the glide-directed and user-directed modes.]{Horizontal box plots showing participants agreement with various statements for the glide-directed and user-directed modes. The statements and agreement percentages are described in the findings section. The statements and the agreement percentages are described in the Findings section.}
  \caption{Participant agreement statement about learnability, ease of use, level of comfort, trust, and confidence for the Glide-directed mode (left) and User-directed mode (right).}
  \label{fig:mode-statements}
\end{figure*}

We recorded whether they completed the course and the time it took them to complete the course. Additionally, we measured the number of errors. This included the number of times the participant became misaligned with Glide (i.e. the user was not standing directly centered behind Glide) and the number of potential collisions (an experimenter intervened before a collision occurred). After the participants completed the course three times, we asked them to state their agreement with a series of statements~(\figref{fig:mode-statements}) and answer the NASA task load index (TLX) on a 7-point Likert scale. Additionally, they answered three open-ended questions: (1) \textit{What did you like most about this mode?} (2) \textit{What did you like least about the mode?} (3) \textit{When would you see yourself using this mode?}

In the second section we evaluated the User-directed mode. We had the user complete three courses of their choosing~(\figref{fig:routes}). There were three possible destinations: lounge, work area, and kitchen. 
At each junction we informed the user which directions they could go to get to various destinations. For example, we would say “Go straight to get to the work area or kitchen or turn left to get to the work area or lounge”. From there, the user could select their desired direction by twisting the handle, receiving the haptic feedback and begin walking. We recorded the same metrics and answers to open-ended questions as we did in the first section.

In the last section participants filled out an exit questionnaire about their experience using the various modes and their recommendations for improving Glide. Note, we did not counterbalance the sections because our goal was not to compare the two modes but to do a general analysis about each individual mode.

\begin{table}[t]
  \caption{Number of errors across participants for each mode.}
  \label{tab:errors}
  \begin{tabular}{llllll}
    \textbf{Mode} & \textbf{Trial} & \textbf{Avg} & \textbf{SD} & \textbf{Min} & \textbf{Max}\\ [-0.7mm]
    \midrule
    \textbf{Glide-directed} & 1 & 3.33 & 3.32 & 0 & 10 \\ [-0.7mm]
    \midrule
    \textbf{Glide-directed} & 2 & 2.25 & 1.39 & 0 & 4 \\ [-0.7mm]
    \midrule
    \textbf{Glide-directed} & 3 & 1.25 & 0.87 & 0 & 3 \\ [-0.7mm]
    \midrule
    \textbf{User-directed} & 1 & 0.44 & 0.73 & 0 & 2 \\ [-0.7mm]
    \midrule
    \textbf{User-directed} & 2 & 0.25 & 0.46 & 0 & 1 \\ [-0.7mm]
    \midrule
    \textbf{User-directed} & 3 & 0.71 & 0.76 & 0 & 2 \\ [-0.7mm]
  \bottomrule
    \end{tabular}
\end{table}

\section{Findings}

\textbf{Trust in Glide}--We define trust as the user's assessment of the reliability of the system. About 70\% of participants trusted Glide to guide them around obstacles~(\figref{fig:mode-statements}) in the Glide-directed mode and approximately 60\% of participants trusted Glide~(\figref{fig:mode-statements}) in the User-directed mode. In general, users trusted Glide as its controller was able to safely steer the user around obstacles and avoid walls in most scenarios. A current limitation of the system is that the user must be centered behind Glide and walk at a slower pace to give the controller enough time and space to steer effectively. If a user was mis-aligned or walking too fast, the controller was sometimes unable to steer users back towards the global plan. Some participants noted that they did not like it when Glide ran into or brushed up against walls; this affected their trust in Glide.

\textbf{Confidence when Using Glide}--We define confidence as the user's assessment in their own abilities to use the system effectively. Approximately 70\% of participants agreed that Glide inspired confidence when they were walking in the Glide-directed mode and a little more than 75\% agreed in the User-directed mode~(\figref{fig:mode-statements}). One participant noted a learning curve between the Glide-directed and User-directed mode, showing an increase in confidence in the User-directed mode: \textit{“The [Glide-directed] mode was mostly learning how to use Glide and I felt more prepared in the [User-directed] mode.”} Additionally, allowing the user to choose which direction they went, by twisting the handle, in the User-directed mode increased confidence: \textit{“The [User-directed] mode made me motivated to move faster than the [Glide-directed] mode. The [Glide-directed] mode made me feel more hesitant because I didn’t know where I was going."} More specifically, at a system level, the twist in the handle processed by the torque sensor and the haptic feedback the system provided to acknowledge the user input increased confidence.

\textbf{Learnability, Ease-of-Use, and Comfortability}--Approximately 95\% of participants agreed that Glide was easy to learn in the Glide-directed mode and 100\% agreed in the User-directed mode~(\figref{fig:mode-statements}). 100\% of participants agreed that Glide was easy to use in the Glide-directed mode and approximately 95\% agreed in the User-directed mode~(\figref{fig:mode-statements}). Participants thought Glide was intuitive and they liked the ease of motion: \textit{“It didn’t present any difficulty in the motion; it didn’t feel forced, it felt natural.”} About 95\% of participants agreed that they could tell when Glide was turning and could follow accordingly in the Glide-directed mode and 100\% agreed in the User-directed mode~(\figref{fig:mode-statements}). Many participants pointed out that they liked the way Glide turned: \textit{“I liked the turning, it was easy for me to tell when I had to turn. The first time I tried it I didn’t even realize it was giving me a signal, I instinctively turned with it.”}, \textit{“I liked that it slowly turned instead of an instant 90-degree angle.”} At at system level, the shape of the global plan influenced the behavior that the participants are describing. The global plan had a large turning radius allowing for more gradual turns. Additionally, approximately 75\% of participants agreed that they felt comfortable with the path Glide set in the Glide-directed mode and 100\% agreed in the User-directed mode~(\figref{fig:mode-statements}). More specifically, the user's ability to choose the direction of the global plans Glide set between junctions in the User-directed mode resulted in increased comfort.

\textbf{Performance}--We measure performance based on the number of errors and the time taken to complete a trial. An error is when the user becomes mis-aligned with Glide resulting in a potential collision. We show the number of errors and trial completion times for both modes across participants in Table 2 and 3. The number of errors across both modes and trial completion times in the Glide-directed mode decreased as the users continued to use Glide. We cannot make a comparison between the trial times in the User-directed mode as users selected routes which were all different lengths. Overall, participants performance improved over time because of their ability to quickly learn how to use Glide for the aforementioned reasons. 

\textbf{Control over Movement}---We designed Glide with passive wheels to give users control over their walking pace. We noticed that some users adjusted their speed based on how Glide was steering. For example, if Glide was turning, some users would walk slower than when walking in a straight line. Additionally, if Glide brushed up against the wall users could stop, back up and reorient Glide. One user in particular was able to sense how close they were to obstacles through echolocation and would slow down if they were close to an obstacle. Unlike a motorized robot, this design allows users to adjust their speed to their liking and also stop and/or reorient if they feel unsafe.

\textbf{Feedback from Glide}--As participants continued to use Glide, they became more comfortable with slowdown haptic and braking. Approximately 75\% of participants agreed that they could easily tell when they had to stop or slowdown in the Glide-directed mode and about 90\% agreed in the User-directed mode~(\figref{fig:mode-statements}). 100\% of the participants agreed that twisting the handle to turn in the User-directed mode made sense to them. Overall, most participants noted that they liked the twisting gesture. One participant said, \textit{“I thought it was intuitive to twist the handle, I didn’t have to put in too much effort or change the orientation of the device. I liked the haptic feedback I got before the brakes, so I knew to slow down.”}

\begin{figure}[t]
  \includegraphics[width=\columnwidth]{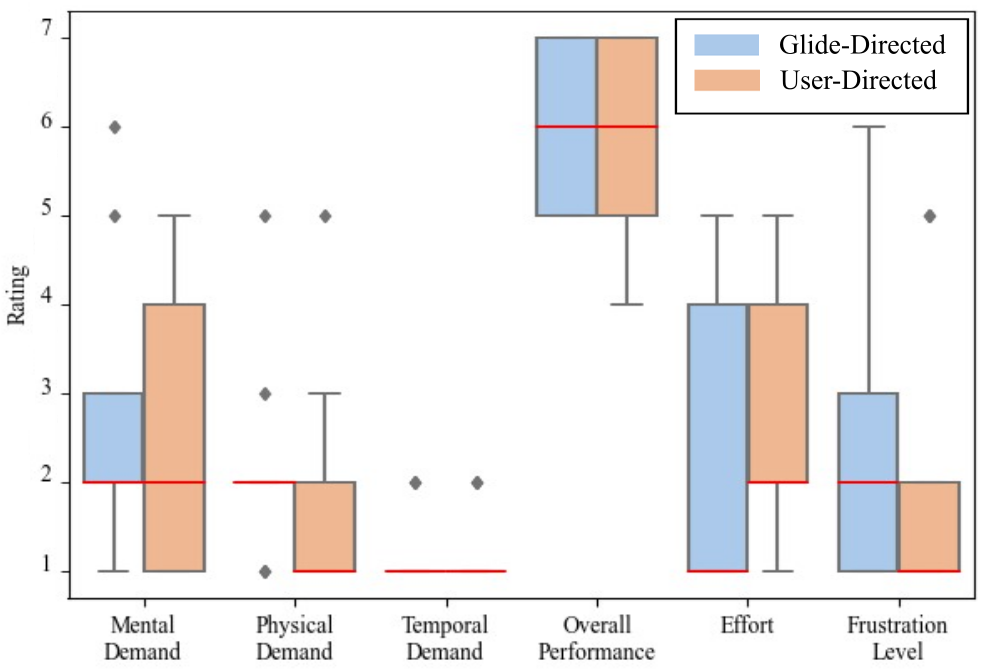}
  \Description[Box plot showing NASA TLX ratings by participants for both modes.]{Box plot showing NASA TLX ratings by participants for both modes. Temporal demand were low for the both modes. Mental demand was low for the Glide-directed mode and a little higher for the user-directed mode. Users rated overall performance as high and effort as medium for both modes. Frustration levels were relatively low for both modes but slightly higher in the Glide-directed mode. }
  \caption{NASA-TLX ratings by participants ranging from (1) Low to (7) High. The red line is the median.}
  \label{fig:tlx}
\end{figure}

\textbf{Task Workload}--The Task load index was assessed after each mode was explored and medians across participants are shown in~\figref{fig:tlx}. Overall, the physical and temporal demand across modes was low and participants thought they performed well with both modes. Participants thought that the User-directed mode was more mentally demanding than the Glide-directed mode: \textit{“The [User-directed] mode was more challenging because I had to make decisions”}, \textit{“I did not have to think in the [Glide-directed] mode but in the [User-directed] mode I had to think more and make more choices.”} 

\textbf{Uses for Glide-directed Mode}--Four participants said they would use the Glide-directed mode in an unknown or crowded area: \textit{“It is useful in crowded unknown situations. Unlike a cane where you collide with obstacles to know where they are, this mode just avoids obstacles for you”}, \textit{“Instead of using assistance to get guided to a conference room in an unfamiliar hotel, Glide could guide me.”}, \textit{“I would use it when I am in a crowded area like a park, festival or farmer’s market.”}, \textit{“I can see myself using it when going into restaurants.”} One participant said that they would never use it and another two participants said that they would use it all the time. One participant suggested combining Glide with a shopping cart so they could push groceries back home. One participant said they were not sure when they would use this mode.

\textbf{Uses for User-directed Mode}--Four participants said they would use the User-directed mode indoors. Two of those participants specifically pointed out they would use it in unfamiliar indoor environments: \textit{“I would use it in unfamiliar environments and if I wanted to specific rooms or locations independently.”}, \textit{“I would use it indoors when receiving verbal feedback of where I can go.”} One participant said that they would use it outdoors when crossing streets. But another participant pointed out that using this mode outdoors would be difficult: \textit{“I generally need to follow google maps when going somewhere. I don’t know how I would manage to hold both Glide and my phone. I think I would need to exert more effort when outdoors.”} One participant said they would use Glide when they wanted to choose the route. Another participant said they would use Glide when they did not want to interact with another person or guide dog.  One participant said they are not sure when they would use this mode.

\section{Limitations \& Future Work}

\textbf{Speed}--We asked user to walk at a slow pace to handle a latency issue with the controller we used. If the user walked too fast the controller was not able to react quickly enough. The users would get too close to a wall and the controller could not find a suitable trajectory to avoid the wall. In these situations, we had to ask the user to stop and back up. Many users commented that they would like to be able to walk faster. 

\textbf{Alignment with Glide}--Participants commented that it was difficult to stay directly behind Glide. One user said, \textit{“I want to try holding the robot next to me instead of in front.”} Glide needs to be able to handle various positions of users with respect to it. Additionally, when the users were not centered behind Glide, they would often veer towards the walls. Glide would compensate by turning the wheels away from the wall, but this would result in a “zig-zagging” motion. 


\textbf{Global Map}--Currently Glide relies on having a global map of the environment. In future iterations we want Glide to be able to map and navigate its environment at the same time. 

\textbf{Complex Environments}--Glide can currently only operate in a single indoor floor plan that is flat. We want Glide to be able to operate in more complex environments with overhangs, stairs, elevators, ramps, etc. Additionally, we want Glide to be able to operate outdoors.

\textbf{Interaction Modalities}--We acknowledge there are some limitations in our work as we did not compare against existing interaction approaches (e.g. audio-based interaction). However, we limited the interaction mechanisms to not overwhelm the users. Despite our efforts, some users still noted that there was a higher mental workload when using the User-directed mode, which has the most interaction capabilities, and preferred the Glide-directed mode~(\figref{fig:tlx}).


\section{Discussion \& Conclusion}
This paper explores various levels of control of assistive navigation that Glide offers. It confirms that Glide is easy to use and easy to learn. Most users trusted Glide but some of the limitations of the current system impacted the remaining users trust in Glide. As users continued to use Glide, they became more confident and their performance improved. 

One key insight is that users’ preferences in modes and their level of autonomy varied. Additionally, users also would use different modes based on the situation they are in. A future direction to explore is customization. It is not possible to design a single mode that encompasses all user preferences. Users should be able to select between modes that they would like to use. 

Furthermore, users had varying opinions about what kinds of interactions and feedback they wanted from Glide. For example, some users said they wanted audio feedback but another user said they would prefer a more complex haptic vocabulary over audio feedback. There is no single interaction technique or type of feedback that is more useful. In future work we plan to experiment with other interaction techniques and feedback and allow the user to select between the various options.

\bibliographystyle{ACM-Reference-Format}
\balance
\bibliography{sample-base}










\end{document}